\UseRawInputEncoding
\documentclass[letterpaper, 10 pt, journal, twoside]{IEEEtran}
\usepackage[utf8]{inputenc}

\IEEEoverridecommandlockouts                              

\usepackage{graphicx}
\usepackage{adjustbox}
\usepackage{amsmath, bm} 
\usepackage{amssymb}
\usepackage{pdfpages}
\usepackage[caption=false]{subfig}  
\usepackage{siunitx}
\usepackage{xcolor}
\usepackage{url}
\usepackage[]{hyperref}
\usepackage{cite}

\newcommand{\fix}[1]{\hypersetup{allcolors=black}{\color{black} {#1}}}
\newcommand{\cfix}[1]{%
  \texorpdfstring{\textcolor{black}{#1}}{#1}%
}
\DeclareRobustCommand{\sectionfix}[1]{%
  \texorpdfstring{\textcolor{black}{#1}}{#1}%
}

\def\BibTeX{{\rm B\kern-.05em{\sc i\kern-.025em b}\kern-.08em
    T\kern-.1667em\lower.7ex\hbox{E}\kern-.125emX}}

\title{MonoSpheres: Large-Scale Monocular SLAM-Based UAV Exploration through Perception-Coupled Mapping and Planning}

\usepackage{scalerel}
\usepackage{tikz}
\usetikzlibrary{svg.path}
\definecolor{orcidlogocol}{HTML}{A6CE39}
\tikzset{
  orcidlogo/.pic={
    \fill[orcidlogocol] svg{M256,128c0,70.7-57.3,128-128,128C57.3,256,0,198.7,0,128C0,57.3,57.3,0,128,0C198.7,0,256,57.3,256,128z};
    \fill[white] svg{M86.3,186.2H70.9V79.1h15.4v48.4V186.2z}
    svg{M108.9,79.1h41.6c39.6,0,57,28.3,57,53.6c0,27.5-21.5,53.6-56.8,53.6h-41.8V79.1z M124.3,172.4h24.5c34.9,0,42.9-26.5,42.9-39.7c0-21.5-13.7-39.7-43.7-39.7h-23.7V172.4z}
    svg{M88.7,56.8c0,5.5-4.5,10.1-10.1,10.1c-5.6,0-10.1-4.6-10.1-10.1c0-5.6,4.5-10.1,10.1-10.1C84.2,46.7,88.7,51.3,88.7,56.8z};
  }
}

\newcommand\orcidicon[1]{\href{https://orcid.org/#1}{\mbox{\scalerel*{
        \begin{tikzpicture}[yscale=-1,transform shape]
          \pic{orcidlogo};
        \end{tikzpicture}
}{|}}}}
\author{
  Tom\'{a}\v{s} Musil$^{\orcidicon{0000-0002-9421-6544}}$, 
  Mat\v{e}j Petrl\'{i}k$^{\orcidicon{0000-0002-5337-9558}}$,
  Martin Saska$^{\orcidicon{0000-0001-7106-3816}}$%
\thanks{%
  Manuscript received November 20, 2025; Revised February 26, 2026; Accepted April 27, 2026.
  This paper was recommended for publication by Editor Pascal Vasseur upon evaluation of the Associate Editor and Reviewers comments.}
  \thanks{This work was funded by the Czech Science Foundation (GAČR) under research project no. 25-17779M, by the European Union under the project Robotics and advanced industrial production (reg. no. CZ.02.01.01/00/22\_008/0004590), and by CTU grant no SGS26/077/OHK3/1T/13.}
  \thanks{%
    Authors are with the Department of Cybernetics, Faculty of Electrical Engineering, Czech Technical University in Prague, 166 36 Prague 6, {\tt\footnotesize\{\href{mailto:musilto8@fel.cvut.cz}{musilto8}|\href{mailto:matej.petrlik@fel.cvut.cz}{matej.petrlik}|\href{mailto:martin.saska@fel.cvut.cz}{martin.saska}\}@fel.cvut.cz}
}
  \thanks{Digital Object Identifier (DOI): see top of this page.}
}
\begin{document}
\maketitle
\markboth{IEEE Robotics and Automation Letters. Preprint Version. Accepted April, 2026}
{Musil \MakeLowercase{\textit{et al.}}: MonoSpheres} 
\begin{abstract}
  Autonomous exploration of unknown environments is a key capability for mobile robots, but it is largely unsolved for robots equipped with only a single monocular camera and no dense range sensors.
  In this paper, we present \fix{MonoSpheres} --- a novel approach to monocular vision-based exploration that can safely cover large-scale unstructured indoor and outdoor 3D environments by explicitly accounting for the properties of a sparse monocular SLAM frontend in both mapping and planning.
  The mapping module solves the problems of sparse depth data, free-space gaps, and large depth uncertainty by oversampling free space in texture-sparse areas and keeping track of obstacle position uncertainty.
  The planning module handles the added free-space uncertainty through rapid replanning and perception-aware heading control.
  We further show that frontier-based exploration is possible with sparse monocular depth data when parallax requirements and the possibility of textureless surfaces are taken into account.
  We evaluate our approach extensively in diverse real-world and simulated environments, including ablation studies.
  To the best of the authors' knowledge, \fix{MonoSpheres} is the first \fix{method} to achieve 3D monocular exploration in real-world unstructured outdoor environments.
  We open-source our implementation 
  to support future research.

  \textit{Code}--- \href{https://github.com/ctu-mrs/monospheres}{github.com/ctu-mrs/monospheres}

  \textit{Videos}--- \href{http://mrs.felk.cvut.cz/monospheres2026}{mrs.felk.cvut.cz/monospheres2026}

\end{abstract}


\section{INTRODUCTION}
\IEEEPARstart{I}{n} recent years, significant advances have been presented in mobile robot autonomy, enabling robot systems to explore unknown 3D environments that span kilometers in scale \cite{darpa_cerberus_wins, csiro_darpa, beneath}. 
These advances are critical for allowing the deployment of uncrewed aerial vehicles (UAVs) in real-world applications, such as search-and-rescue, wildfire response or planetary exploration.
However, the mapping and planning methods used in unknown environments currently rely on dense range sensors, such as LiDARs or depth cameras.
Such sensors are heavy, expensive, and thus severely limit the affordability and flight time of autonomous UAVs.

{Monocular cameras and inertial measurement units (IMUs) are orders of magnitude lighter and less expensive, but monocular-only exploration of unknown environments has only been demonstrated at the scale of a few indoor rooms \cite{from_monoslam_to_explo, los_maps, simon2023mononav, cnn_explo_singleroom}.
The main challenge can be attributed to the fact that the existing methods depend on \textit{reliable} and \textit{dense} depth estimates and obtaining such estimates from a monocular camera is not yet fully solved \cite{deep_mono_depth_2021, monodepth_survey_big_2024}.
Thus, when classical mapping methods are used with any existing monocular depth estimation method, the resulting maps are either too sparse to allow planning in large-scale environments \cite{from_monoslam_to_explo, los_maps}, or they contain wrongly estimated free space, which leads to crashes \cite{simon2023mononav, mono_airsim_nav_2025}.
}

\begin{figure}[!t]
  \centering
\begin{minipage}{0.995\linewidth}
  \begin{tikzpicture}
    \node[anchor=south west,inner sep=0] at (0,0) {%
      \adjincludegraphics[width=\linewidth,
        trim={{0.0\width} {0.1\height} {0.0\width} {0.1\height}}, clip=true]{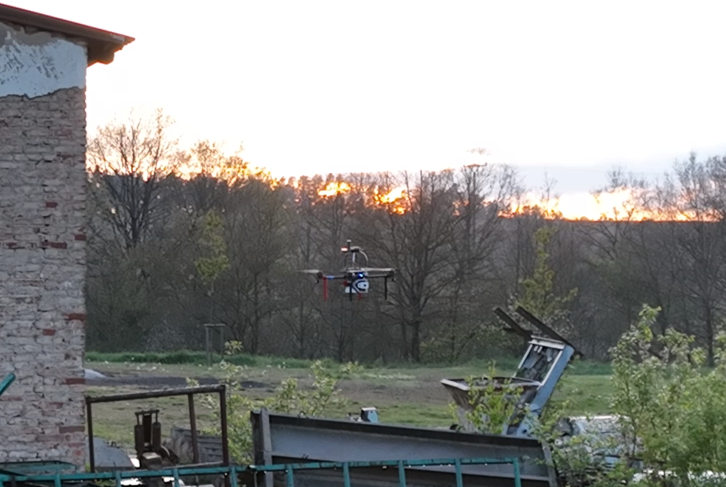}};
  \end{tikzpicture}
\end{minipage}

\vspace{0.2em} 

\begin{minipage}{0.49\linewidth}
  \begin{tikzpicture}
    \node[anchor=south west,inner sep=0] at (0,0) {%
      \adjincludegraphics[width=\linewidth,
        trim={{0.0\width} {0.0\height} {0.0\width} {0.0\height}}, clip=true]{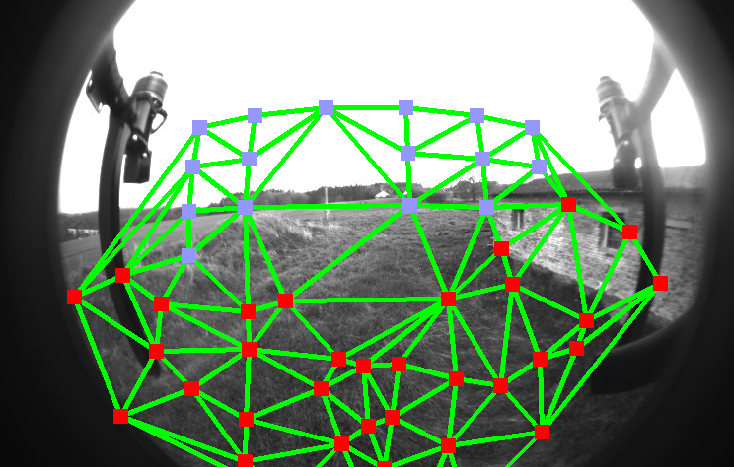}};
  \end{tikzpicture}
\end{minipage}\hfill
\begin{minipage}{0.49\linewidth}
  \begin{tikzpicture}
    \node[anchor=south west,inner sep=0] at (0,0) {%
      \adjincludegraphics[width=\linewidth,
        trim={{0.0\width} {0.0\height} {0.0\width} {0.0\height}}, clip=true]{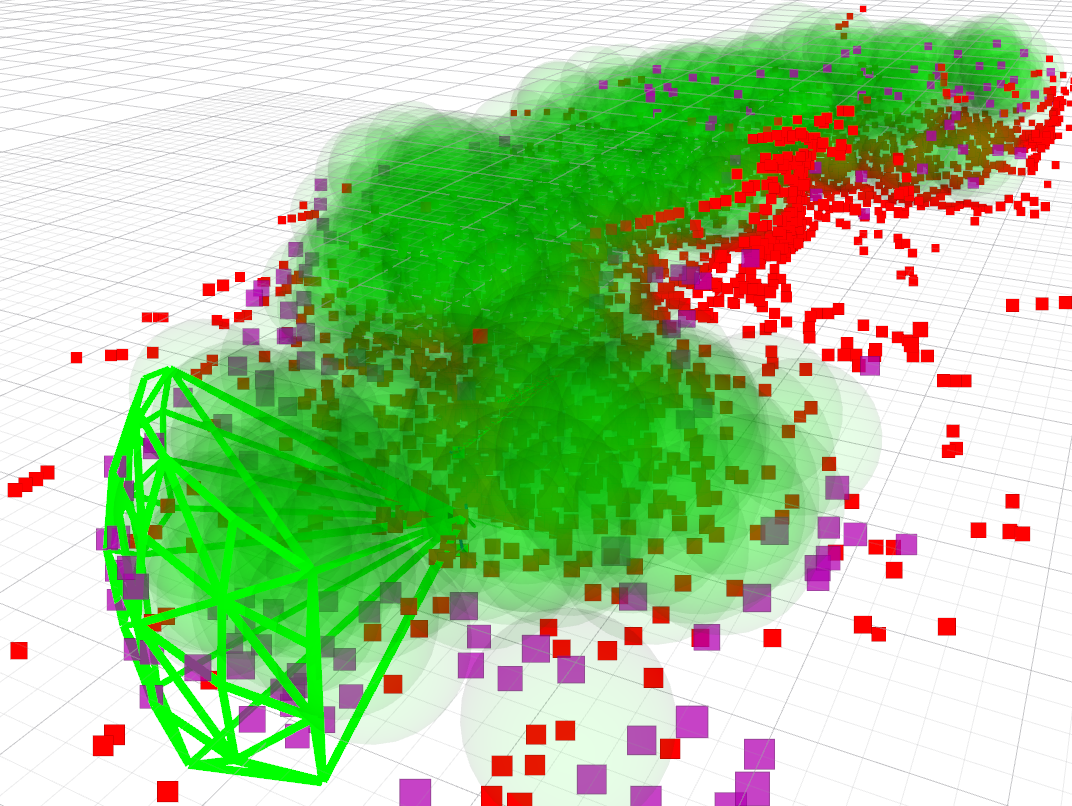}};
  \end{tikzpicture}
\end{minipage}

  \caption{Illustration of the proposed approach. The mapping pipeline {estimates depth using virtual (blue) and tracked (red) points} from sparse monocular-inertial SLAM (left) running on board the UAV (top) to construct a large-scale map (right) consisting of free-space spheres (green), obstacle points (red) {and frontier points (magenta).}}
  \label{fig:smap_intro}
\end{figure}

{In this paper, we identify three core properties of motion-based depth sensing and show that explicitly designing the mapping and planning based on these principles can enable autonomy both indoor and outdoor, and at scales comparable to dense sensing methods.
In essence, our mapping approach interpolates depth and intelligently estimates free space in texture-sparse areas, solving the problem of sparse free space, while keeping track of obstacle position uncertainties to avoid wrongly deleting mapped obstacles.
Our planning approach handles the uncertain depth estimates by flying in the direction of the camera, rapidly replanning when new information is obtained and ensuring pure translational motion when visiting exploration viewpoints.

We evaluate our \fix{combined mapping and planning approach, dubbed MonoSpheres,} in multiple real-world scenarios, simulations, and ablation studies, demonstrating that it 
achieves exploration in both indoor and outdoor environments, and at considerably larger scales than existing monocular exploration methods.
As a step towards low-cost vision-based UAV autonomy in unknown environments, our paper brings the following contributions:
}

\begin{itemize}
  \item {A novel monocular 3D mapping method with perception-coupled modules that significantly increase the reliability of mapped free space and obstacles with unreliable, sparse, motion-based depth data as input.}
  \item A perception-aware exploration approach, enabled by the proposed mapping method, that achieves large-scale 3D outdoor exploration on a UAV equipped with only a monocular camera and IMU, 
      validated in real-world conditions.
    \item Open-sourced code for the proposed methods, along with example simulation scripts for replicating our results, providing a benchmark for new monocular exploration methods.
\end{itemize}

\section{Problem Statement}
{
The problem addressed in this paper is actively building an explicit 3D map of free and occupied space in a bounded unknown environment.
In the case of robots equipped with dense depth sensors (most commonly LiDAR, RGBD, stereo cameras), this is a well-studied research problem \cite{darpa_cerberus_wins, beneath, paper_frontier_grandpa}.
These works implicitly assume that from any given viewpoint, the sensor will uncover all obstacles up to its sensing range, and that non-detection of obstacles means that free-space can be initialized in the given direction  \cite{occupancy_moravec, octomap, voxblox, ufomap}.

However, this assumption does not apply for monocular depth sensing.
Motion-based depth sensing, as used in monocular SLAM algorithms, can only estimate depth for textured surfaces and requires sufficient parallax to do so.
Single-image learning-based monocular depth estimation methods \cite{deep_mono_depth_2021, monodepth_survey_big_2024} can provide dense depth estimates, but they are not yet reliable enough for safe navigation in general unknown environments \cite{simon2023mononav, mono_airsim_nav_2025}.
Thankfully, motion-based depth sensing does have some regularities that can be exploited to design mapping and planning that could handle the uncertain and sparse depth estimates.
We base our solution to this problem on the following key assumptions/observations about monocular motion-based depth sensing applicable to most of the monocular SLAM algorithms:
\begin{enumerate}
  \item {Depth accuracy generally increases with decreasing distance to obstacles}, as the \textit{inverse} of depth errors are gaussian \cite{inverse_depth} and since objects occupy more pixels in the image, making them more likely to be detected.
  \item In outdoor environments, {the UAV can encounter large textureless areas where no depth can be estimated by classic motion-based methods} (e.g. above horizon on a clear day) that initialize depth by tracking texture.
  \item {Measuring depth requires sufficient translational motion}, making the depth measurements \textit{trajectory dependent}, which is not the case with dense range sensors.
\end{enumerate}
}

\section{Related Works}
\label{sec:related_works}

{One approach to solving the monocular exploration problem is to use deep learning-based monocular depth estimation models to essentially turn the monocular camera into a dense depth sensor, and then use classical mapping and planning methods.
The authors of \cite{simon2023mononav, cnn_explo_singleroom} have demonstrated using learning-based depth estimation models combined with traditional navigation approaches.
However, the authors of \cite{simon2023mononav} demonstrate autonomy only at the scale of a single room, while also encountering collisions and map deformations due to depth estimation errors.
The authors of \cite{mono_airsim_nav_2025} present navigation with obstacle avoidance using pretrained and finetuned monocular depth models in larger outdoor environments in simulation, but also note a high collision and navigation failure rate due to the same problems.
As further discussed in recent surveys \cite{deep_mono_depth_2021, monodepth_survey_big_2024}, current depth estimation models still struggle to provide real-time onboard performance and reliable generalization in domains that the model was not trained on.
In addition, they require a GPU which significantly raises the cost and weight of a UAV.

The other main approach to monocular exploration is to use sparse 3D points estimated by visual SLAM as the only depth measurements and work with those.
Most of the research on using motion-based depth estimation has been on reactive obstacle avoidance \cite{flame, avoidance_mono1} with no map building.
Some works already investigated exploration with motion-based depth sensing, but they make limiting assumptions on the environment's structure (e.g. corridors only \cite{corridors_mono_nav_2009} or considering a single object with no other obstacles \cite{information_driven_bbx_mapping}).

For general monocular exploration of \textit{unstructured} environments, the authors of \cite{from_monoslam_to_explo, los_maps} investigated building an occupancy grid \cite{occupancy_moravec} by tracing rays of free space towards low-covariance SLAM points, similarly to classic mapping with dense range sensors. 
As documented in \cite{from_monoslam_to_explo}, this approach causes gaps to appear in the free space of the map when the depth measurements are sparse in low-texture environments.
The authors of \cite{from_monoslam_to_explo} further present a perception-aware exploration method specifically to cover these gaps and demonstrate an increase in mapped volume.
However, the authors show this exploration approach to only work at the scale of a single room.
The authors of \cite{los_maps} also trace rays towards sparse SLAM points, and they present exploration on the scale of multiple indoor rooms in simulation.
They work around the gap issue by setting the map voxel size to be relatively large compared to the UAV size.
A larger voxel size causes a higher percentage of voxels to be passed by the sparse rays towards measured points, reducing the gaps, but can cause narrow passages to be blocked in the map if they are not perfectly aligned with the voxel grid.

\begin{figure*}[!t]
\centering
\begin{tikzpicture}
  \node[anchor=south west,inner sep=0] (a) at (0,0) {
    \adjincludegraphics[width=0.99\linewidth,
      trim={{0.1\width} {0.00\height} {0.05\width} {0.05\height}}, clip=true]{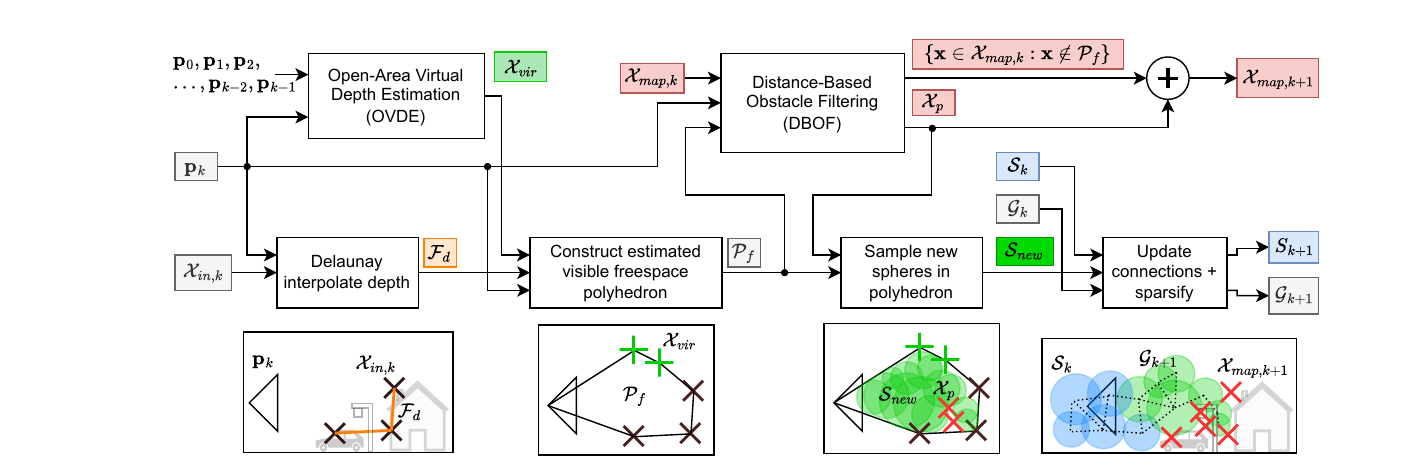}};
\end{tikzpicture}
  \caption{
    {Mapping pipeline overview. The OVDE and DBOF modules are illustrated in \autoref{fig:ofs} and \autoref{fig:distbased} respectively.}
  }
  \label{fig:mapping_pipeline}
\end{figure*}

Based on the findings of the above described works, we decided that monocular depth sensing is so different from dense sensors, that not only the planning, but also the \textit{mapping} should be redesigned to explicitly take into account the specific depth estimation properties.
Since safety and scalable representation of both large-scale and narrow spaces are our main concerns, we chose to use a graph-of-spheres representation \cite{spheregraph, spheremap, bubbleplanner} as the world model instead of voxel grids \cite{octomap, ufomap} or TSDFs \cite{voxblox} traditionally used in exploration literature.
In comparison to the existing sphere-based mapping methods however, our proposed approach is the first to build the map \textit{directly} from \textit{sparse} monocular depth data, as the others require a full voxel-based occupancy map to be built with dense sensors to extract the graph of spheres. 
Compared to existing monocular exploration methods overall, \fix{MonoSpheres} achieves exploration at larger scales, both indoor and outdoor, in unstructured 3D environments and without needing a GPU, thanks to the proposed perception-coupled mapping and planning modules.
}

\section{Sphere-Based Mapping Using Sparse Visual SLAM Points}
\label{sec:map_building}
{The proposed mapping method represents free space by a graph $\mathcal{G}$ of intersecting spheres $\mathcal{S}$ and obstacles as a set of 3D points $\mathcal{X}_{map}$.}
Each sphere at the $k$-th update iteration has a static center $\mathbf{c}$ and changing radius $r_k$.
The radius $r_k$ represents the distance from $\mathbf{c}$ to the nearest unknown space or obstacle at time step $k$.
{To allow rapid path planning, we maintain a graph $\mathcal{G}$, which contains an edge for each pair of intersecting spheres.
To represent obstacles, we accumulate the stable 3D points (i.e. with depth covariance below a fixed threshold $\gamma$) estimated by a sparse monocular SLAM frontend into a set of points $\mathcal{X}_{map}$ with a minimal point-to-point distance $\xi$.
The obstacle points are updated, merged and removed based on criteria designed specifically for monocular sensing, described in \autoref{sec:distance-based}.
A single map update iteration consists of the following steps, visualized also in \autoref{fig:mapping_pipeline}.
}
\subsection{Depth Interpolation and Polyhedron Construction}
\label{sec:polyhedron}
{We interpolate depth between the sparse SLAM points using a simplified version of the approach described in FLAME \cite{flame}, where
a precise mesh is constructed from visual keypoint measurements.
In this paper, we care primarily about mapping the free space for navigation purposes, and thus we do not perform the mesh optimization or split the 2D interpolations as in \cite{flame}.
We project the currently tracked visual SLAM points onto the image plane and compute their Delaunay triangulation. 
By connecting the points in 3D according to their Delaunay triangulation in 2D, we obtain a mesh $\mathcal{F}_d$ that we call a \textit{depth mesh}, which is used only for the current frame.
We estimate currently visible free space as the volume between the camera's position $\mathbf{p}$ and all points on $\mathcal{F}_d$.
This space is enclosed by connecting all points that lie on the edge of $\mathcal{F}_d$ to the camera's focal point, which forms an \textit{estimated visible free space polyhedron} $\mathcal{P}_f$.
It is only used for the update step at the current time and discarded afterwards.
}

\begin{figure}[!h]
      \begin{tikzpicture}
        \centering
        \node[anchor=south west,inner sep=0] (a) at (0,0) {\adjincludegraphics[width=1.0\linewidth, trim={{0.0\width} {0.05\height} {0.0\width} {0.15\height}}, clip=true]{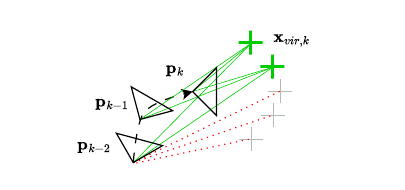}};
      \end{tikzpicture}
  \centering
  \caption{{Open-area virtual-depth estimation diagram. The thick green points $\mathbf{x}_{vir, k}$ satisfy the condition defined in \autoref{sec:fake} and are added to the construction of $\mathcal{F}_d$\cfix{.}}}
  \label{fig:ofs}
\end{figure}

\subsection{Open-Area Virtual-Depth Estimation (OVDE)}
\label{sec:fake}
{When moving in open areas, such as a grassy field without tall obstacles in \autoref{fig:smap_intro}, point-based visual SLAM can accurately localize only nearby points corresponding to the ground. 
Using the ray-tracing grid-based mapping method \cite{from_monoslam_to_explo, los_maps}, this would cause free space to be estimated only in a downward direction and not allow flight across the field.
To allow flight in open areas, our method estimates additional free space based on the following assumption: 
Consider a virtual point in space $\mathbf{x}_{vir}$ that falls into the camera's FoV at the current pose $\mathbf{p}_k$ and also at previous poses $\mathbf{p}_{k-n}, ...,\mathbf{p}_{k-2}, \mathbf{p}_{k-1}$.
If there is sufficient parallax between any two of these poses for estimating the distance of $\mathbf{x}_{vir}$ 
(based on the max depth covariance of the measured SLAM points $\gamma$), 
and if the visual SLAM frontend has not detected any obstacle point approximately in the direction of  $\mathbf{x}_{vir}$, then $\mathbf{x}_{vir}$ must lie in free space (see \autoref{fig:ofs}).

The mapping method periodically checks 
the described condition for a set of virtual points $\mathbf{x}_{vir}$ at predefined fixed positions relative to the UAV (see \autoref{fig:ofs}) at each mapping iteration and for a fixed number of previous poses. 
The points that fulfill this condition are used to recompute the \textit{depth mesh} $\mathcal{F}_d$ as if they were measured by the SLAM frontend.
However, we discard points that would project into the original Delaunay triangulation of the current SLAM points, since there is a more reliable depth estimate from interpolating the SLAM points.
Free space estimated in this manner also cannot erase existing obstacle points in the map.
\fix{This approach allows flexibly adding more free space in open areas while using the SLAM points for estimating free space near obstacles.}

}

\begin{figure}[!h]
  \includegraphics[width=8.5cm]{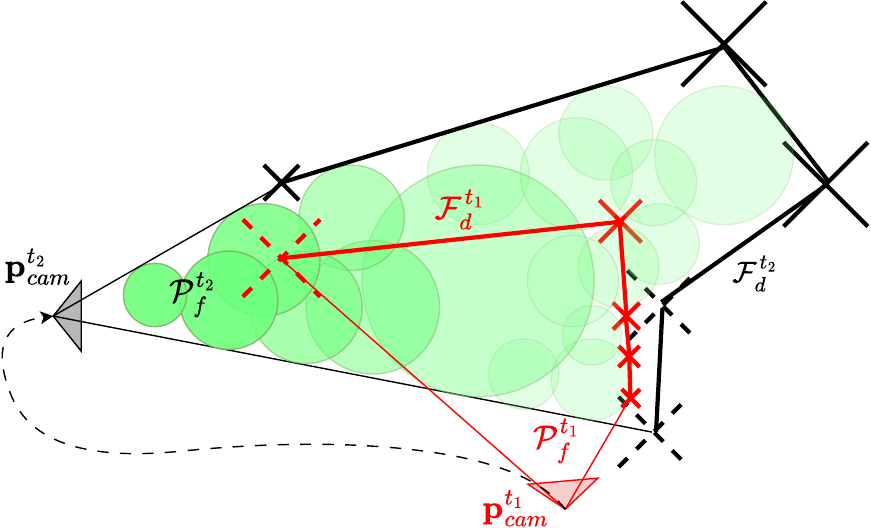}
  \centering
  \caption{Top-down illustration of sphere sampling and obstacle point management: 
  The visible free-space polyhedron $\mathcal{P}_f^{t_2}$ is created using the {low-covariance SLAM points tracked at $t_2$} (black crosses) and the camera's {pose} $\mathbf{p}_{cam}^{t_2}$.
  The size of a point $\mathbf{x}$ {corresponds} to the distance $d_{\mathbf{x}}$ that it was measured from.
  {Since our method prioritizes closer measurements (of obstacles and lack thereof), only the (dashed red) point seen far away from the camera at a previous time $t_1$, which now lies in $\mathcal{P}_f^{t_2}$ very close to the camera, will be deleted as noise.}
  New spheres are inscribed inside $\mathcal{P}_f^{t_2}$, but their radii are also bound by the protected points (solid red) $\mathcal{X}_p$, which {were added to the map at $t_1$ and not deleted at $t_2$ even though they lie in $\mathcal{P}_f^{t_2}$.
  The (dashed black) two points observed far from the camera at $t_2$ are not added to the map, as there are more precisely localized (red) points nearby.} }
  \label{fig:distbased}
\end{figure}

\subsection{Distance-Based Obstacle Filtering (DBOF)}
\label{sec:distance-based}
{As the next step of the update, we decide which tracked SLAM points to add into the map points $\mathcal{X}_{map}$, and which points to delete from the map.
An important piece of information is the position covariance of each SLAM point.
Since in inverse depth parametrization, the position covariance of a point grows with the distance from which it was measured \cite{inverse_depth}, we approximate the covariance simply as the lowest distance $d_{m, x}$ that a point has been observed from.}

{The main issue addressed in this module is that, as shown in \autoref{fig:mapping_pipeline} and \autoref{fig:distbased}, points corresponding to a small obstacle might momentarily not be detected by the SLAM frontend, while a larger obstacle behind the small one is detected.}
Thus, we cannot simply delete every map point that falls into $\mathcal{P}_f$, because then the original points would be wrongly deleted, {potentially causing collisions}.
We solve this problem by deleting any map point $\mathbf{x}$ that falls into $\mathcal{P}_f$ only if
\begin{equation}
  \mathbf{x} \in \mathcal{P}_f \quad \text{and}  \quad |\mathbf{x} - \mathbf{p}_{cam}| < \alpha \cdot d_{m,x},
\end{equation}
where $\mathbf{p}_{cam}$ is the position of the camera and $\alpha$ is a parameter, which we empirically set to $1.1$ to account for distance measurement noise.
The points that lie in the free-space polyhedron $\mathcal{P}_f$ and are not deleted from the map are labeled as \textit{protected points} $\mathcal{X}_p$.
These points are used to constrain sphere radii in the following update step.

{Additionally, to prioritize more accurate depth measurements, points seen from a closer distance replace points seen from larger distances, if they are measured closer than $\xi$ to them.} 
In the same way, new points seen at a large distance are not added to the map, if there are more accurately localized points near them, which is also visualized in \autoref{fig:distbased}. 
We enforce this rule by deleting any newly added or previously mapped point $\mathbf{x}$ if
\begin{equation}
  \exists \mathbf{x}_{prev}: |\mathbf{x} - \mathbf{x}_{prev}| < \xi \quad \text{and} \quad d_{m,x_{prev}}  < d_{m,x}
\end{equation}
where $\mathbf{x}_{prev}$ is another mapped or newly added point with a higher position accuracy and {$\xi$ is the obstacle map resolution.}

\subsection{Updating and Adding New Spheres}
\label{sec:existing_spheres_update}
Next, we recompute the radii of existing spheres that could be updated by $\mathcal{P}_f$ or the input points $\mathcal{X}_{in}$.
To bound the update time of this step, we check the largest sphere radius $r_{max}$ in the map, which allows us to quickly filter out all spheres whose centers fall outside a bounding box around $\mathcal{P}_f$, inflated by $r_{max}$. 
Then, for any remaining sphere with a center $\mathbf{c}$ and radius $r_{k}$, the updated radius is computed as
\begin{equation}
  r_{k+1} = \min \left( \max \left( r_{k}, d(\mathbf{c}, \mathcal{P}_f) \right)
  , d(\mathbf{c}, \mathcal{X}_{in} \cup \mathcal{X}_p) \right).
  \label{eq:update}
\end{equation}
In this equation, $d(\mathbf{c}, \mathcal{P}_f)$ is the signed distance to $\mathcal{P}_f$ (positive if the point is inside the polyhedron, negative if outside).
The value of $d(\mathbf{c},\mathcal{X}_{in} \cup \mathcal{X}_p)$ is the minimum distance to all input obstacle points $\mathcal{X}_{in}$, and to the protected map points $\mathcal{X}_p$ described in \autoref{sec:distance-based}.
Essentially, this equation means that the polyhedron $\mathcal{P}_f$ is used to increase the radius of a sphere, whereas the protected $\mathcal{X}_p$ and input points $\mathcal{X}_{in}$ are used to constrain it.
As the last part of this step, we delete all spheres with $r_{k+1} < r_{min}$, where $r_{min}$ is the smallest allowed sphere radius specified by the user.

{To introduce new spheres into the map, we sample a fixed number of points inside $\mathcal{P}_f$ in uniformly sampled directions at uniformly sampled distances between the camera and the depth mesh $\mathcal{F}_d$.}
A potential new sphere's radius is determined in the same way as for the old spheres in the previous step in Eq. \ref{eq:update} with $r_k = 0$.
If the potential radius is larger than $r_{min}$, the sphere is added to the map.

\subsection{Recomputing and Sparsifying Sphere Graph}
\label{sec:graph_update}
After all the sphere radii updates have been made, we update the graph of spheres $\mathcal{G}$ used for path planning, so that all intersecting spheres are connected in the graph.
Furthermore, to constrain map update time and path planning time, we perform a redundancy check on the updated and added spheres in the same way as in \cite{spheremap}.
If any sphere is found to be redundant, it is deleted from the map.
This step is important to cover large open areas by only a few spheres to allow rapid planning, and tight corridors by a higher density of spheres, capturing the information about potential paths and distances in more detail.

{\section{Perception-Coupled Exploration with a Sparse Sphere-Based Map}
{In this section, we describe the proposed perception-coupled navigation and exploration approach, visualized in \autoref{fig:exploration_explanation}.}
In principle, we employ the commonly used greedy next-best-view (NBV) \cite{paper_frontier_grandpa} strategy, which has so far been deemed unsuitable for monocular exploration in previous works \cite{from_monoslam_to_explo, los_maps}.
In the following sections, we introduce several major differences in methodology that are critical for allowing the NBV strategy to be used for robots equipped with only a monocular camera for depth sensing.

\subsection{Frontier Sampling on Free-Space Polyhedron}
Firstly, since the proposed map representation is fundamentally different from an occupancy grid, we need to detect frontiers (i.e. the boundary of free and unknown space) in a different way than with an occupancy grid.
{We propose to sample points along the visible free-space polyhedron $ \mathcal{P}_f $  described in \autoref{sec:map_building} at each map update, and add the points as frontiers, if they do not lie inside any free-space sphere and if they are at a pre-defined distance $\zeta$ from all map obstacle points and other frontiers.
If they do not meet these criteria in any following update, they are deleted. 
To avoid the loss of SLAM tracking due to exploration of completely textureless areas, we also delete frontiers that are further than $\zeta$ from any obstacle points.}
These frontier points are used to generate exploration viewpoints such that at least some frontier points are visible from each viewpoint.

\begin{figure}[!htb]
 \def\subfigwidth{0.48\linewidth}
  \centering

  \subfloat{%
    \begin{minipage}{\subfigwidth}
      \begin{tikzpicture}
        \node[anchor=south west,inner sep=0] (a) at (0,0)
          {\adjincludegraphics[width=1.0\linewidth,
            trim={{0.08\width} {0.1\height} {0.08\width} {0.1\height}},
            clip=true]{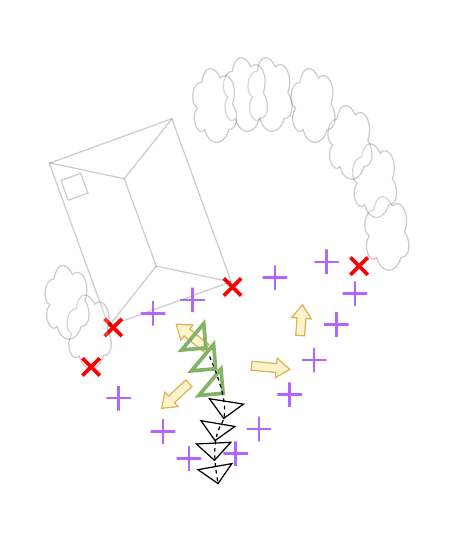}};
        \node[anchor=south west, font=\bfseries, text=black,
              xshift=2pt, yshift=2pt] at (a.south west) {(a)};
      \end{tikzpicture}
    \end{minipage}
  }%
  \hfill
  \subfloat{%
    \begin{minipage}{\subfigwidth}
      \begin{tikzpicture}
        \node[anchor=south west,inner sep=0] (b) at (0,0)
          {\adjincludegraphics[width=1.0\linewidth,
            trim={{0.08\width} {0.1\height} {0.08\width} {0.1\height}},
            clip=true]{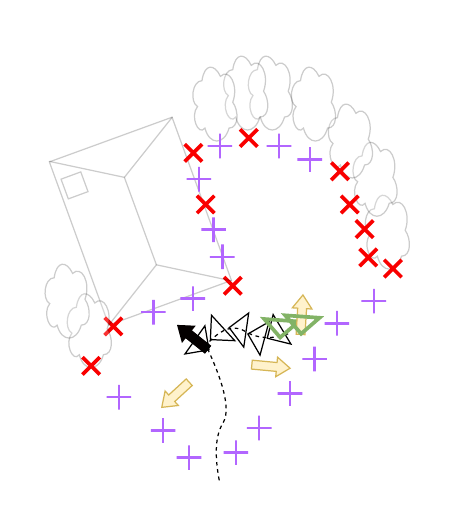}};
        \node[anchor=south west, font=\bfseries, text=black,
              xshift=2pt, yshift=2pt] at (b.south west) {(b)};
      \end{tikzpicture}
    \end{minipage}
  }%
     \hfill
  \caption{
    {Top-down illustration of the exploration approach.
    In (a), the UAV navigates to one of the sampled viewpoints (yellow) aimed at a feature-sparse wall.
    This does not uncover any frontiers (magenta), so to stop returning to this or nearby viewpoints, the viewpoint is blocked (black).
    In both (a) and (b), the UAV flies facing forward and aligns itself with the viewpoint's direction when nearing it (green poses).
    }
    }
  \label{fig:exploration_explanation}
\end{figure}

{\subsection{Perception-Aware Heading Control}}
\label{sec:align}
In traditional exploration approaches with dense distance sensors, it is sufficient to move the robot to a frontier, aim the sensors towards the unknown space, and assume that the distance sensors will uncover some additional space behind the frontier and thus expand the map.
This approach will not, in principle, work reliably with a robot using a monocular camera for estimating depth from motion.
Such a robot requires \textit{translational} motion to obtain correct distance estimates of visual keypoints. 
With rotation only, or motions that have too much rotation compared to translation, reliable motion-based depth estimates cannot be obtained.

To solve this, we propose a simple perception-aware planning approach: 
When finding a path to an exploration  viewpoint, we force the trajectory to aim the UAV in the viewpoint's direction after getting to a pre-defined distance $d_c$ from the goal, as visualized in \autoref{fig:exploration_explanation}.
This way, the UAV reaches the viewpoint with at least $d_c$ meters of {purely translational motion}.
If textured surfaces are visible from that viewpoint, the UAV will most likely observe sufficient parallax to estimate their distances.

{When not aligning itself with a viewpoint as described above,
the UAV is controlled to always aim in the direction of its movement.
This approach is simple, but as we show in \autoref{sec:ablations}, it is crucial for ensuring collision-free flight in the case of incorrectly initialized free space.
Using the forward-facing flight (FFF), the UAV has a higher chance to see obstacles that were not detected from afar, and quickly replan.
To further increase safety and allow agile maneuvers, path planning on the graph of spheres uses the criterion combining path length and distances to obstacles (sphere radii) described in \cite{spheremap}, making it prefer slightly longer paths with higher clearance from obstacles where possible.
}

\subsection{Explored Viewpoint Blocking}
\label{sec:blocking}
Another {important} distinction we propose for NBV exploration with a monocular camera is to block sampling of new exploration viewpoints near visited viewpoints.
{This is done because with motion-based depth sensing, no points will ever be measured on textureless surfaces.}
Frontier points will be generated there instead, since the surfaces lie on the edge of the visible free space polyhedron and can be far enough from any obstacle points.
However, such frontiers cannot uncover any new space, since there is nothing behind them.
For this reason, we block the sampling of new viewpoints near visited viewpoints, {visualized as black arrows in \autoref{fig:exploration_explanation} and \autoref{fig:exper_real}.}
This stops the UAV from getting stuck repeatedly trying to uncover such surfaces.

\section{Experiments}
\label{sec:experiments}
In this section, we analyze the performance of \fix{MonoSpheres} in large-scale real-world (Sec. \ref{sec:real_exper}) and simulated (Sec. \ref{sec:sim_exper}) environments.
The presented implementation of the proposed mapping and exploration methods {is currently written in python in two threads.}
{With this implementation, the mapping runs on average at $\SI{5}{\hertz}$ on 
an AMD Ryzen 7 4800 CPU.
The second loop -- frontier and goal viewpoint updates and long-distance path planning -- runs at approx. $\SI{4}{\hertz}$ in the evaluated worlds.
This is sufficient for exploration at moderate speeds, but can be slow to react to obstacles at high speeds.
For this reason, the UAV also checks the predicted trajectory for potential collisions with the input pointclouds at the SLAM's update rate and stops on collision detection.
The UAV used in the experiments was equipped with a monocular global-shutter grayscale fisheye Bluefox camera, and ICM-42688 IMU.
The MRS UAV system \cite{mrs_uav_system} was used for low-level motion planning, control, and the grid-based path planning for the simulation comparisons.
For state estimation and as the source of sparse visual SLAM points, we used OpenVINS \cite{openvins}.
}

\begin{figure}[!htb]
\centering
  \begin{tikzpicture}
    \node[anchor=south west,inner sep=0] (imgA) at (0,0) {
      \adjincludegraphics[width=0.49\linewidth,
        trim={{0.0\width} {0.0\height} {0.0\width} {0.0\height}},
        clip=true]{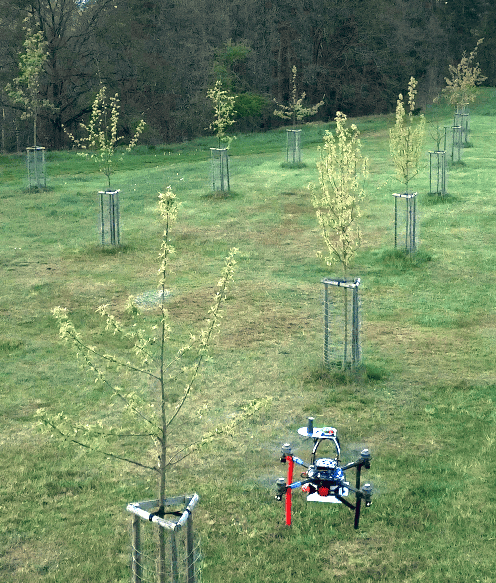}
    };

    \node[anchor=south west,inner sep=0] (imgB) at ([xshift=0.51\linewidth]imgA.south west) {
      \begin{tikzpicture}
        \node[anchor=south west,inner sep=0] (image) at (0,0) {
          \adjincludegraphics[width=0.49\linewidth,
            trim={{0.1\width} {0.06\height} {0.1\width} {0.13\height}},
            clip=true]{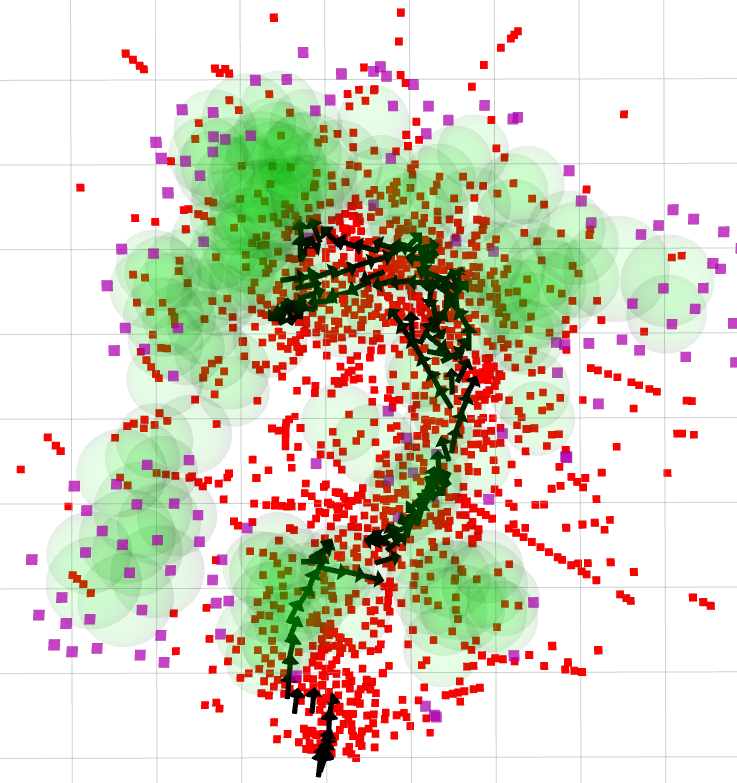}
        };
        \begin{scope}[x={(image.south east)}, y={(image.north west)}]
          \draw[blue, very thick] (0.45,0.35) circle (10pt);
          \draw[blue, very thick] (0.68,0.34) circle (10pt);
          \draw[blue, very thick] (0.7,0.55) circle (10pt);
          \draw[blue, very thick] (0.72,0.78) circle (10pt);
          \draw[blue, very thick] (0.47,0.57) circle (10pt);
          \draw[blue, very thick] (0.48,0.8) circle (10pt);
          \draw[blue, very thick] (0.43,0.12) circle (10pt);
          \draw[blue, very thick] (0.68,0.10) circle (10pt);
        \end{scope}
      \end{tikzpicture}
    };

    \draw[->, ultra thick, blue]
      ([yshift=-0.7cm, xshift=1.5cm]imgA.west) -- ([yshift=1.6cm, xshift=-0.65cm]imgB.south);
    \draw[->, ultra thick, blue]
      ([yshift=0.4cm, xshift=3cm]imgA.west) -- ([yshift=2.7cm, xshift=-0.6cm]imgB.south);
  \end{tikzpicture}
  
  \caption{
    {The 4min orchard exploration experiment. Right: explored map (approx. 40x30m wide), the UAV's trajectory (small black arrows) and trees (circled blue).}
  }
  \label{fig:orchard}
\end{figure}
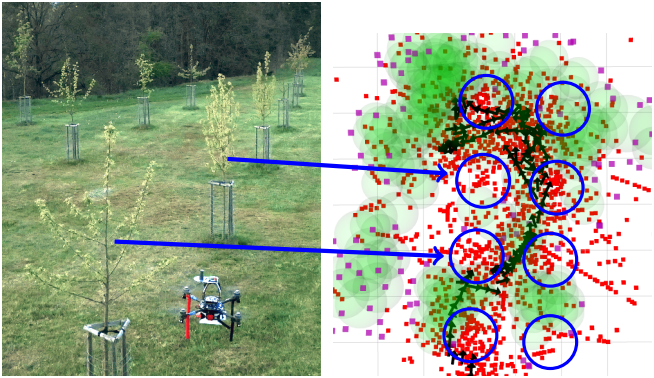

\subsection{Real-World UAV Monocular-Inertial Exploration}
\label{sec:real_exper}
In \autoref{fig:exper_real}, we present an \SI{8}{\minute} experiment where our approach was validated in real-world conditions.
We bounded the area for generating exploration goals to a 70x10x8 $\SI{}{\meter}$ box around the side of an abandoned farmhouse, and the UAV fully autonomously explored nearly all the available space in this region.
The UAV successfully avoided and mapped all the debris, bushes and the house walls in the area, and when battery levels were getting low, it autonomously returned to the starting position.
Also note that the UAV mapped a considerable amount of space in the open field to the left of the house thanks to our approach to safe estimation of free space in open areas, described in \autoref{sec:fake}.
{We have also successfully validated our approach, with the same parameters, in a smaller, more cluttered orchard environment with thin tree branches, shown in \autoref{fig:orchard}.}
Videos from these experiments are available in the multimedia materials. 

\begin{figure}[htb]
\def\subfigwidth{0.49\linewidth}
\centering

\begin{minipage}{\subfigwidth}
  \begin{tikzpicture}
    \node[anchor=south west,inner sep=0] (a) at (0,0) {
      \adjincludegraphics[width=1.0\linewidth,
        trim={{0.0\width} {0.0\height} {0.0\width} {0.0\height}}, clip=true]{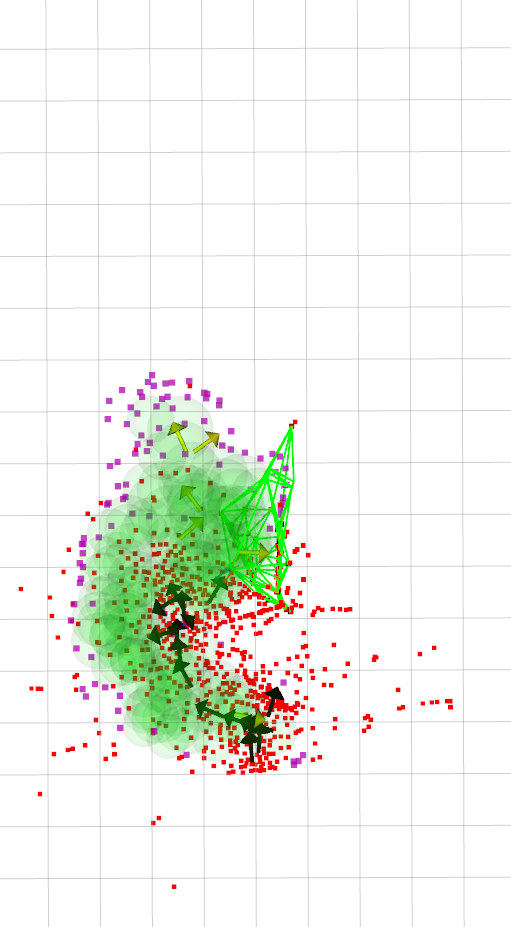}};
    \begin{scope}[x={(a.south east)},y={(a.north west)}]
      \node[align=center] at (0.2, 0.925) {\footnotesize \color{black}$t=126\text{s}$};
    \end{scope}
  \end{tikzpicture}
\end{minipage}%
\hfill
\begin{minipage}{\subfigwidth}
  \begin{tikzpicture}
    \node[anchor=south west,inner sep=0] (a) at (0,0) {
      \adjincludegraphics[width=1.0\linewidth,
        trim={{0.0\width} {0.0\height} {0.0\width} {0.0\height}}, clip=true]{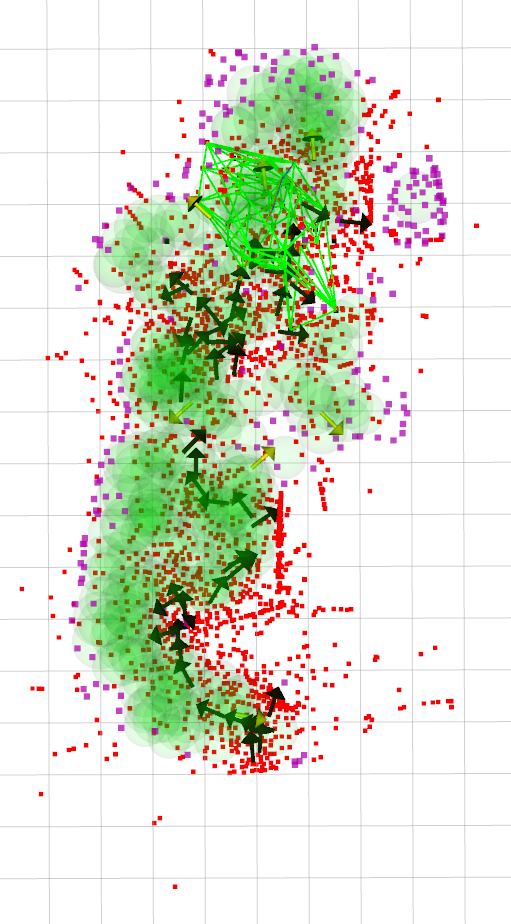}};
    \begin{scope}[x={(a.south east)},y={(a.north west)}]
      \node[align=center] at (0.2, 0.925) {\footnotesize \color{black}$t=370\text{s}$};
    \end{scope}
  \end{tikzpicture}
\end{minipage}

\begin{minipage}{\subfigwidth}
  \begin{tikzpicture}
    \node[anchor=south west,inner sep=0] (a) at (0,0) {
      \adjincludegraphics[width=1.0\linewidth,
        trim={{0.0\width} {0.0\height} {0.0\width} {0.0\height}}, clip=true]{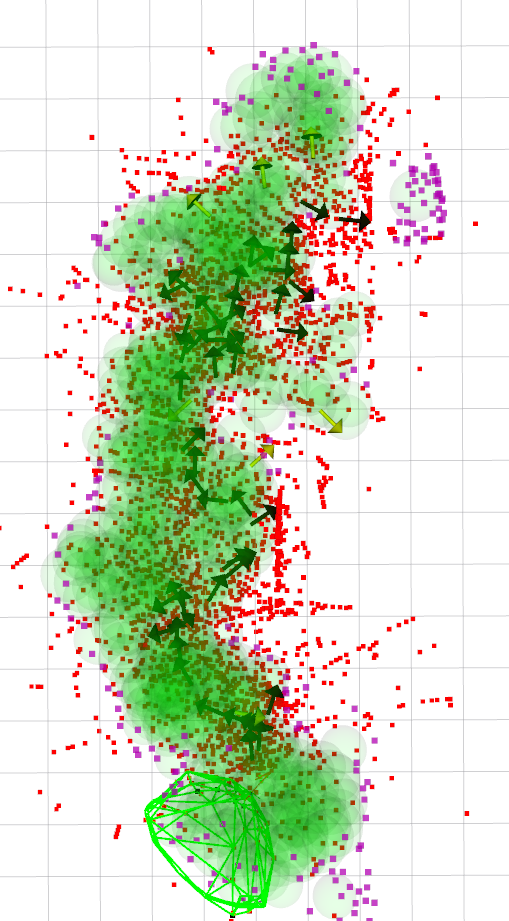}};
    \draw [latex-latex](4.05,0.4) -- (4.05,7.3);
    \node[align=center] at (3.6, 3.4) {\scriptsize \color{black}\SI{80}{\meter}};
    \begin{scope}[x={(a.south east)},y={(a.north west)}]
      \node[align=center] at (0.2, 0.925) {\footnotesize \color{black}$t=516\text{s}$};
    \end{scope}
  \end{tikzpicture}
\end{minipage}%
\hfill
\begin{minipage}{\subfigwidth}
  \begin{tikzpicture}
    \node[anchor=south west,inner sep=0] (a) at (0,0) {
      \adjincludegraphics[angle=180, origin=c, width=1.0\linewidth,
        trim={{0.15\width} {0.34\height} {0.2\width} {0.0\height}}, clip=true]{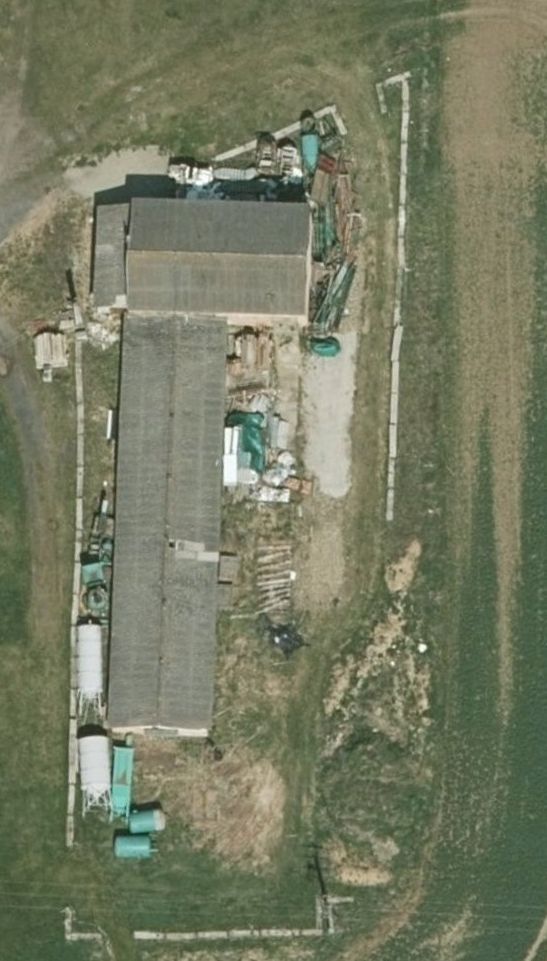}};
  \end{tikzpicture}
\end{minipage}
  \caption{
    Visualization of the real-world large-scale exploration experiment described in \autoref{sec:real_exper}, along with a satellite image of the explored abandoned farm area. Free space (green) is shown alongside mapped obstacle points (red), frontiers (purple) and explored viewpoints (black). 
  }
  \label{fig:exper_real}
\end{figure}
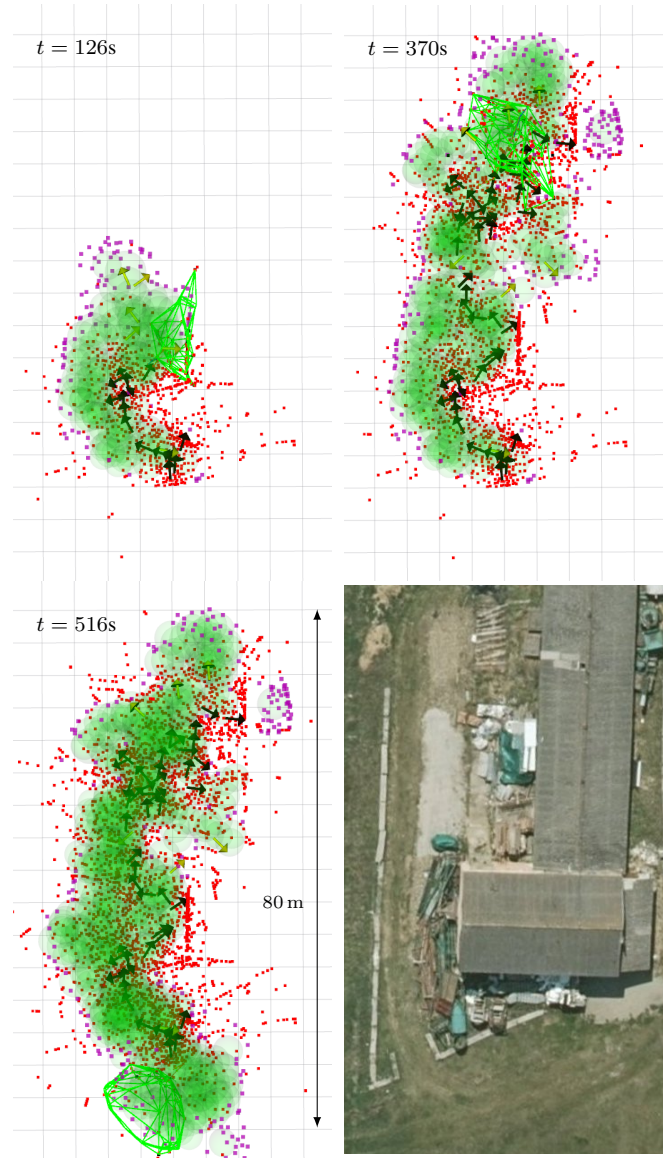

\newcommand{\markX}[3][0.01]{%
  \draw[line width=2.8pt, black]
    (#2-#1, #3-#1) -- (#2+#1, #3+#1);
  \draw[line width=2.8pt, black]
    (#2-#1, #3+#1) -- (#2+#1, #3-#1);
  \pgfmathsetmacro{\innerscale}{0.85}
  \pgfmathsetmacro{\s}{#1*\innerscale}
    \draw[line width=1.2pt, white]
      (#2-\s, #3-\s) -- (#2+\s, #3+\s);
    \draw[line width=1.2pt, white]
      (#2-\s, #3+\s) -- (#2+\s, #3-\s);
}

\newcommand{\textbox}[2]{%
  \node[
    anchor=south east,                        
    inner sep=2pt,                             
    fill=white!30,                             
    draw=black,                                
    font=\sffamily\fontsize{9}{9}\selectfont,  
    text=black,
    line width=0.8pt
  ] at ([xshift=-3pt,yshift=3pt]#1.south east) {#2};  
}

\begin{figure*}[!htb]
\def\subfigwidth{0.24\linewidth}
\centering

\begin{minipage}{\subfigwidth}
  \begin{tikzpicture}
    \node[anchor=south west,inner sep=0] (a) at (0,0) {
      \adjincludegraphics[width=1.0\linewidth,
        trim={{0.03\width} {0.0\height} {0.03\width} {0.0\height}}, clip=true]{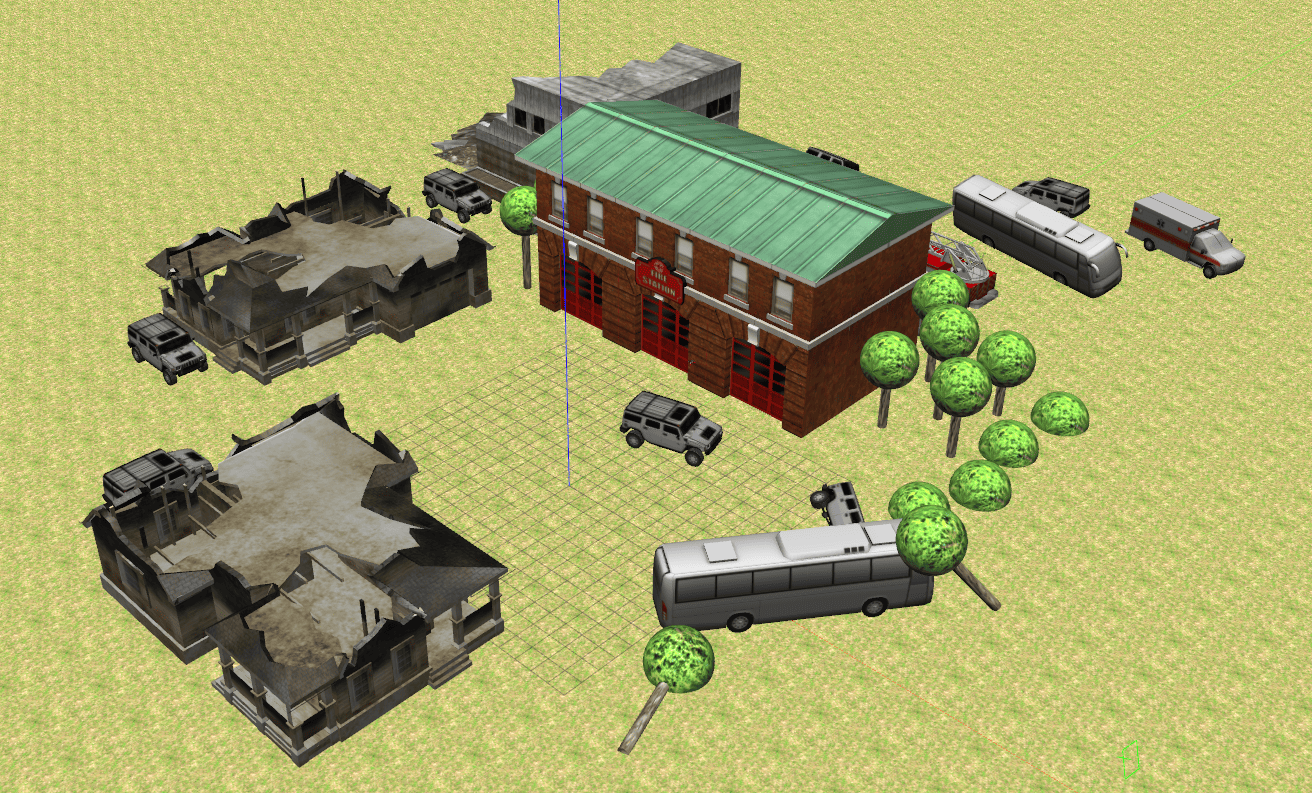}};
    \markX[0.15]{1.8}{1.2}
    \textbox{a}{Earthquake}
  \end{tikzpicture}
\end{minipage}\hfill
\begin{minipage}{\subfigwidth}
  \begin{tikzpicture}
    \node[anchor=south west,inner sep=0] (a) at (0,0) {
      \adjincludegraphics[width=1.0\linewidth,
        trim={{0.08\width} {0.0\height} {0.1\width} {0.0\height}}, clip=true]{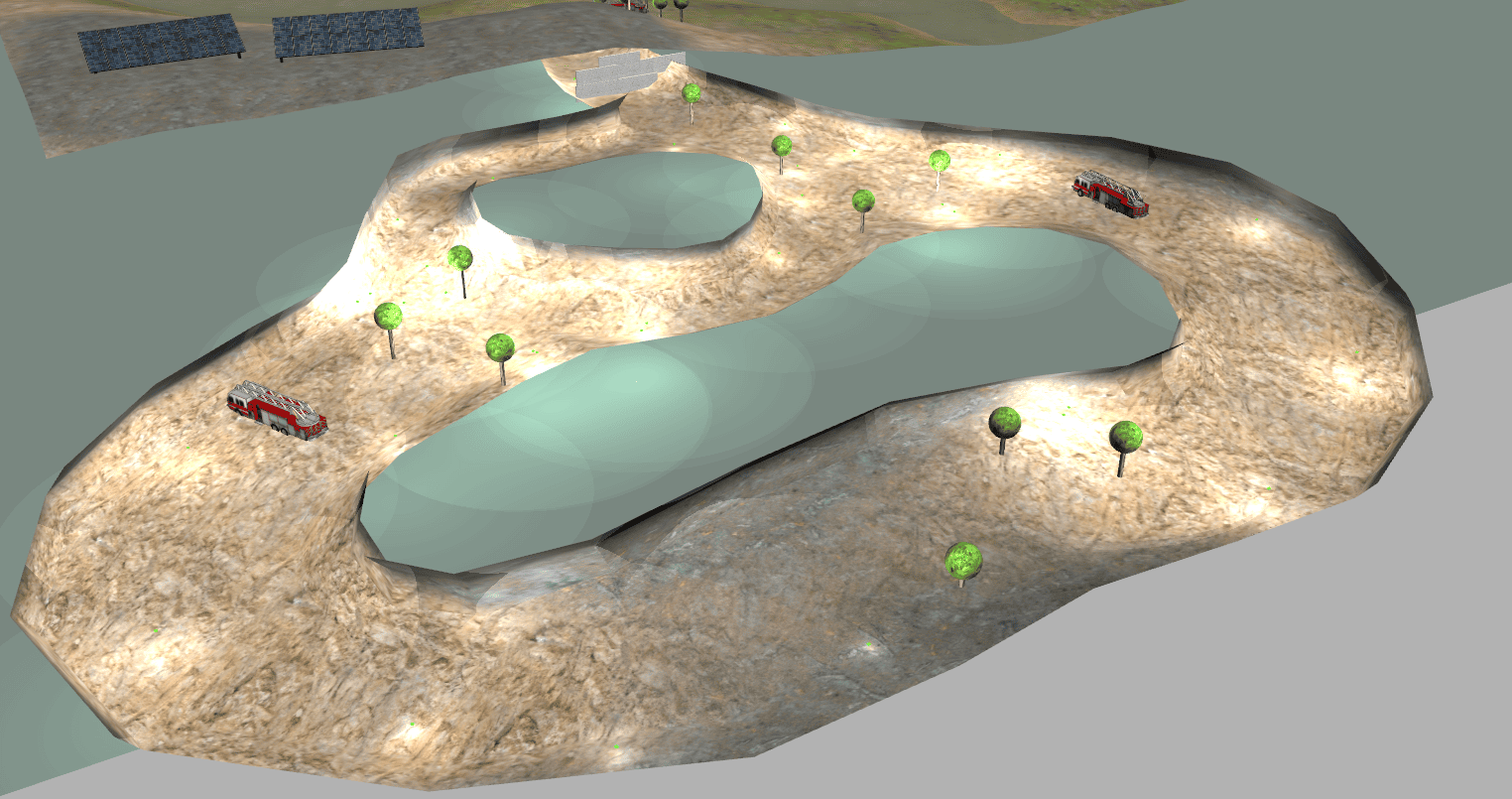}};
    \markX[0.15]{2.6}{2.2}
    \textbox{a}{Cave}
  \end{tikzpicture}
\end{minipage}\hfill
\begin{minipage}{\subfigwidth}
  \begin{tikzpicture}
    \node[anchor=south west,inner sep=0] (a) at (0,0) {
      \adjincludegraphics[width=1.0\linewidth,
        trim={{0.0\width} {0.0\height} {0.0\width} {0.05\height}}, clip=true]{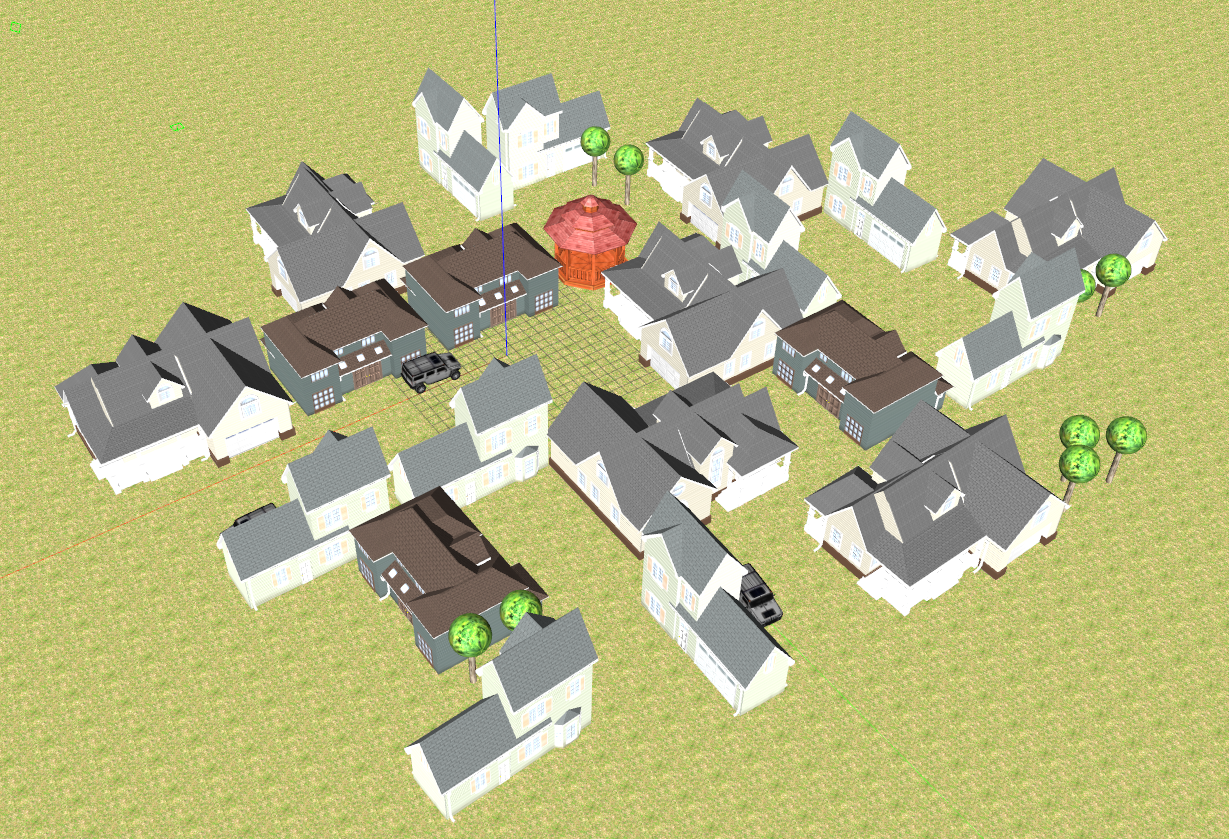}};
    \markX[0.15]{1.9}{1.7}
    \textbox{a}{Rooftops}
  \end{tikzpicture}
\end{minipage}\hfill
\begin{minipage}{\subfigwidth}
  \begin{tikzpicture}
    \node[anchor=south west,inner sep=0] (a) at (0,0) {
      \adjincludegraphics[width=1.0\linewidth,
        trim={{0.0\width} {0.0\height} {0.0\width} {0\height}}, clip=true]{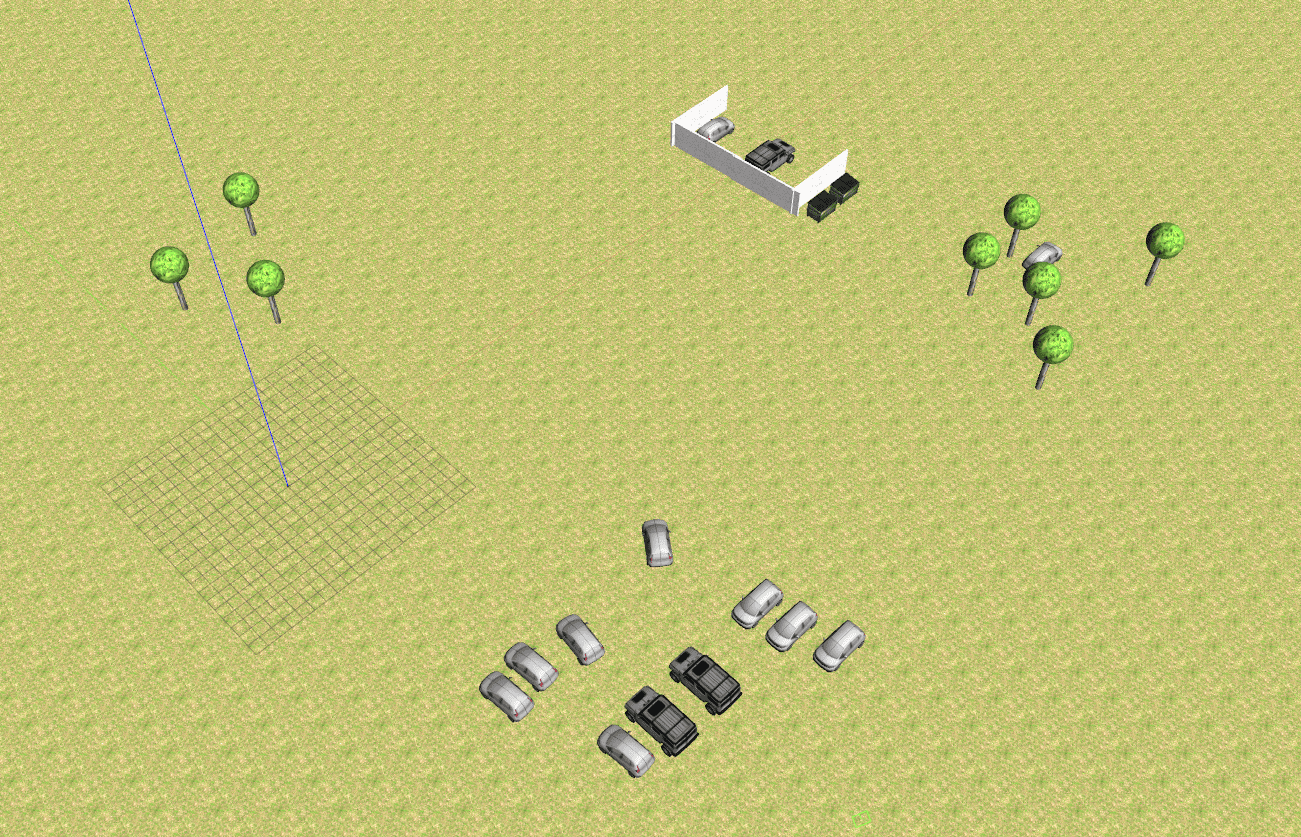}};
    \markX[0.15]{0.95}{1.2}
    \textbox{a}{Sparse}
  \end{tikzpicture}
\end{minipage}

\begin{minipage}{\subfigwidth}
  \begin{tikzpicture}
    \node[anchor=south west,inner sep=0] (a) at (0,0) {
      \adjincludegraphics[width=1.0\linewidth,
        trim={{0.0\width} {0.0\height} {0.0\width} {0.0\height}}, clip=true]{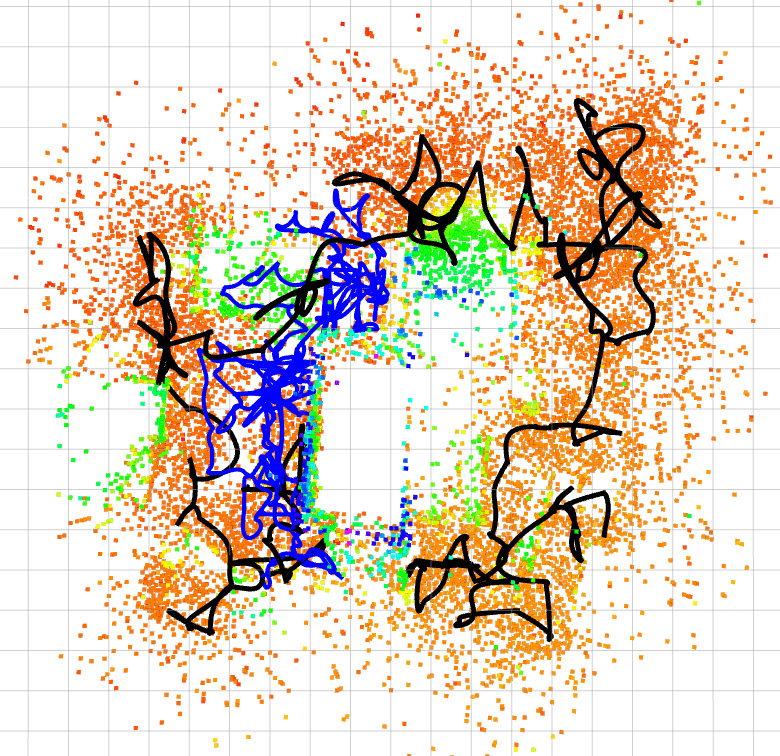}};
    \markX[0.15]{1.4}{1.5}
  \end{tikzpicture}
\end{minipage}\hfill
\begin{minipage}{\subfigwidth}
  \begin{tikzpicture}
    \node[anchor=south west,inner sep=0] (a) at (0,0) {
      \adjincludegraphics[width=1.0\linewidth,
        trim={{0.0\width} {0.0\height} {0.0\width} {0.0\height}}, clip=true]{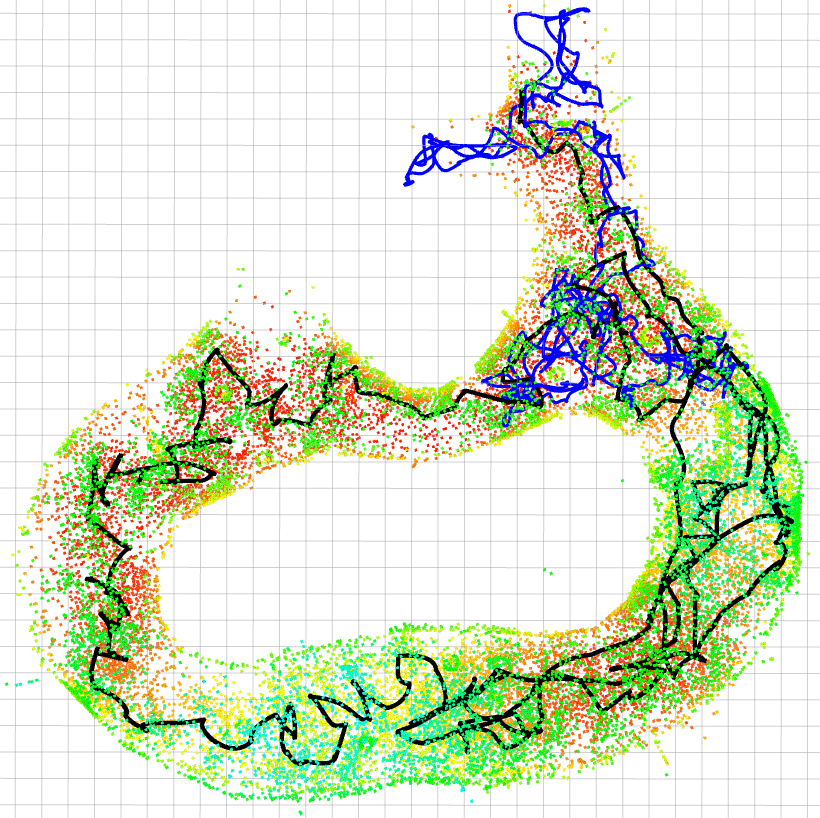}};
    \markX[0.15]{3.2}{2.6}
  \end{tikzpicture}
\end{minipage}\hfill
\begin{minipage}{\subfigwidth}
  \begin{tikzpicture}
    \node[anchor=south west,inner sep=0] (a) at (0,0) {
      \adjincludegraphics[width=1.0\linewidth,
        trim={{0.0\width} {0.0\height} {0.0\width} {0.0\height}}, clip=true]{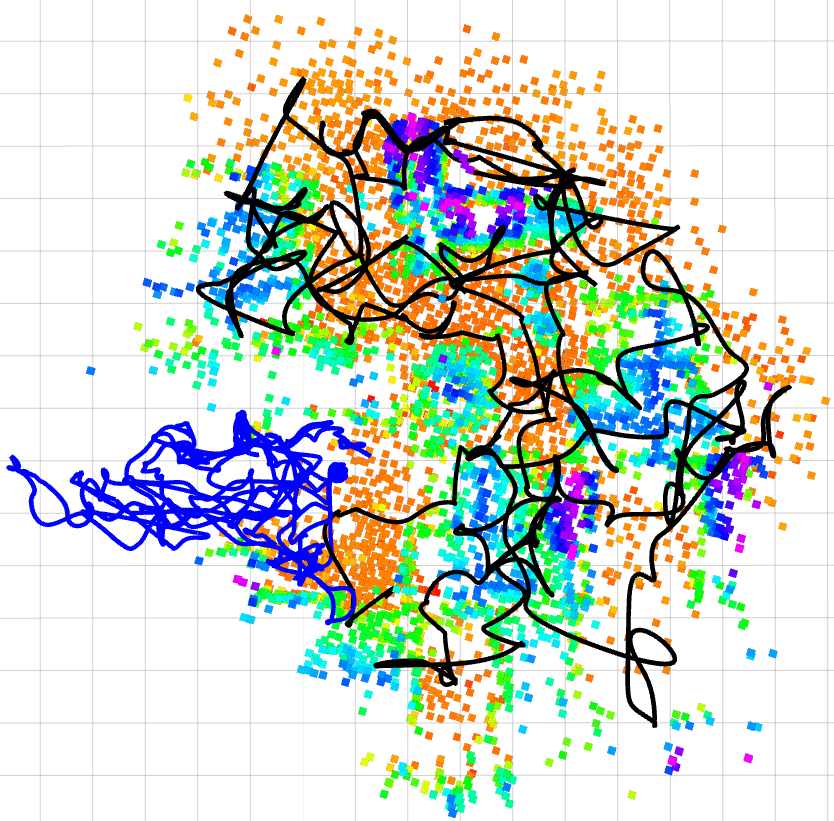}};
    \markX[0.15]{1.7}{1.6}
  \end{tikzpicture}
\end{minipage}\hfill
\begin{minipage}{\subfigwidth}
  \begin{tikzpicture}
    \node[anchor=south west,inner sep=0] (a) at (0,0) {
      \adjincludegraphics[width=1.0\linewidth,
        trim={{0.0\width} {0.0\height} {0.0\width} {0.0\height}}, clip=true]{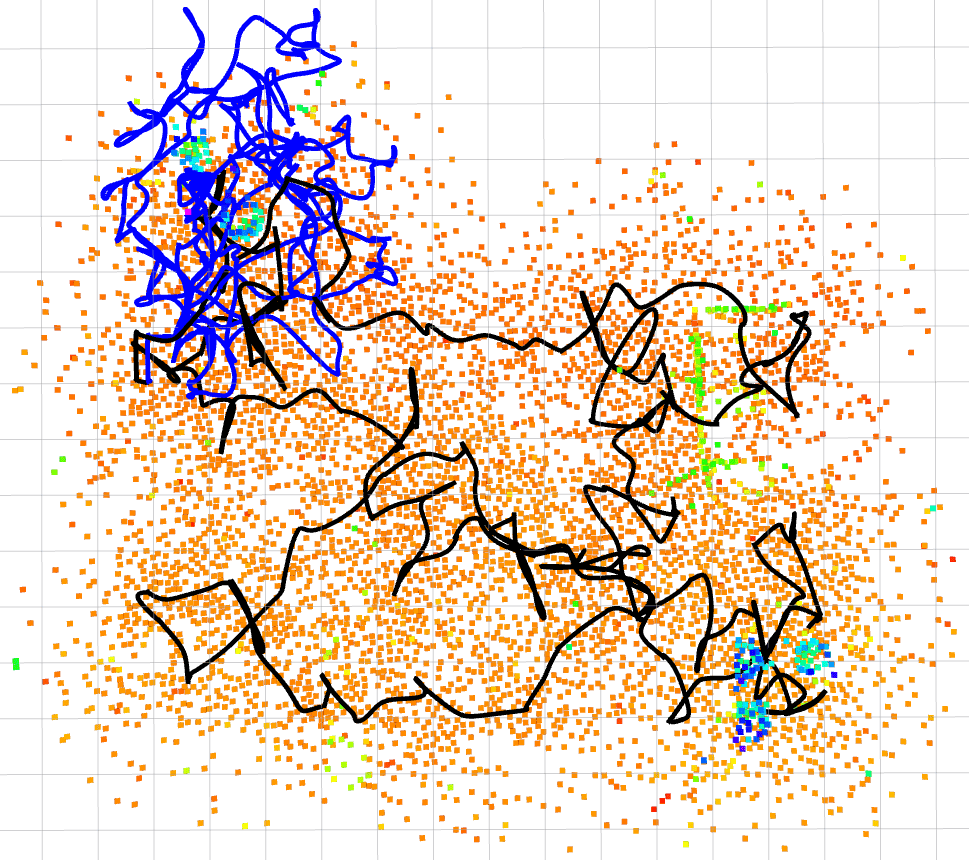}};
    \markX[0.15]{0.95}{1.8}
  \end{tikzpicture}
\end{minipage}

  \centering
  \caption{{Trajectories of the best MonoSpheres (black) and Grid-Based explorer (blue) simulation experiments per each of the 4 worlds (top) along with the resulting MonoSpheres mapped obstacle points (bottom) colored by height. X marks the starting position.}
  }
  \label{fig:exper_fireworld}
\end{figure*}

\begin{figure}[htb]
\def\subfigwidth{0.49\linewidth}
\centering

\begin{minipage}{\subfigwidth}
  \begin{tikzpicture}
    \node[anchor=south west,inner sep=0] (a) at (0,0) {
      \adjincludegraphics[width=1.0\linewidth,
        trim={{0.0\width} {0.0\height} {0.07\width} {0.0\height}}, clip=true]{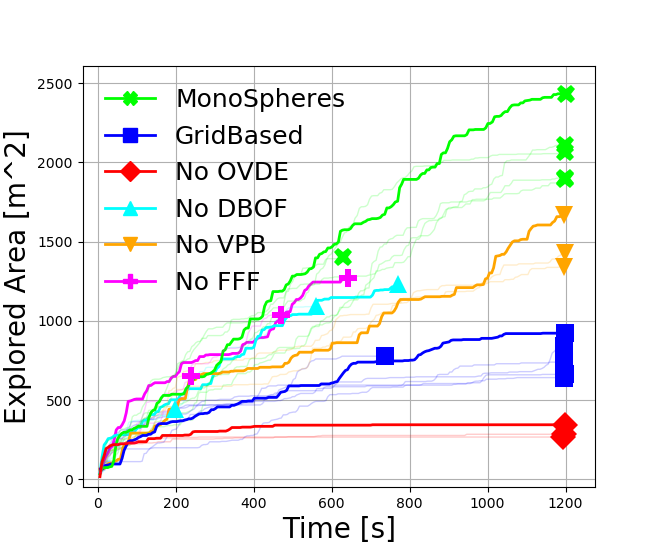}};
  \end{tikzpicture}
\end{minipage}\hfill
\begin{minipage}{\subfigwidth}
  \begin{tikzpicture}
    \node[anchor=south west,inner sep=0] (a) at (0,0) {
      \adjincludegraphics[width=1.0\linewidth,
        trim={{0.15\width} {0.0\height} {0.15\width} {0.0\height}}, clip=true]{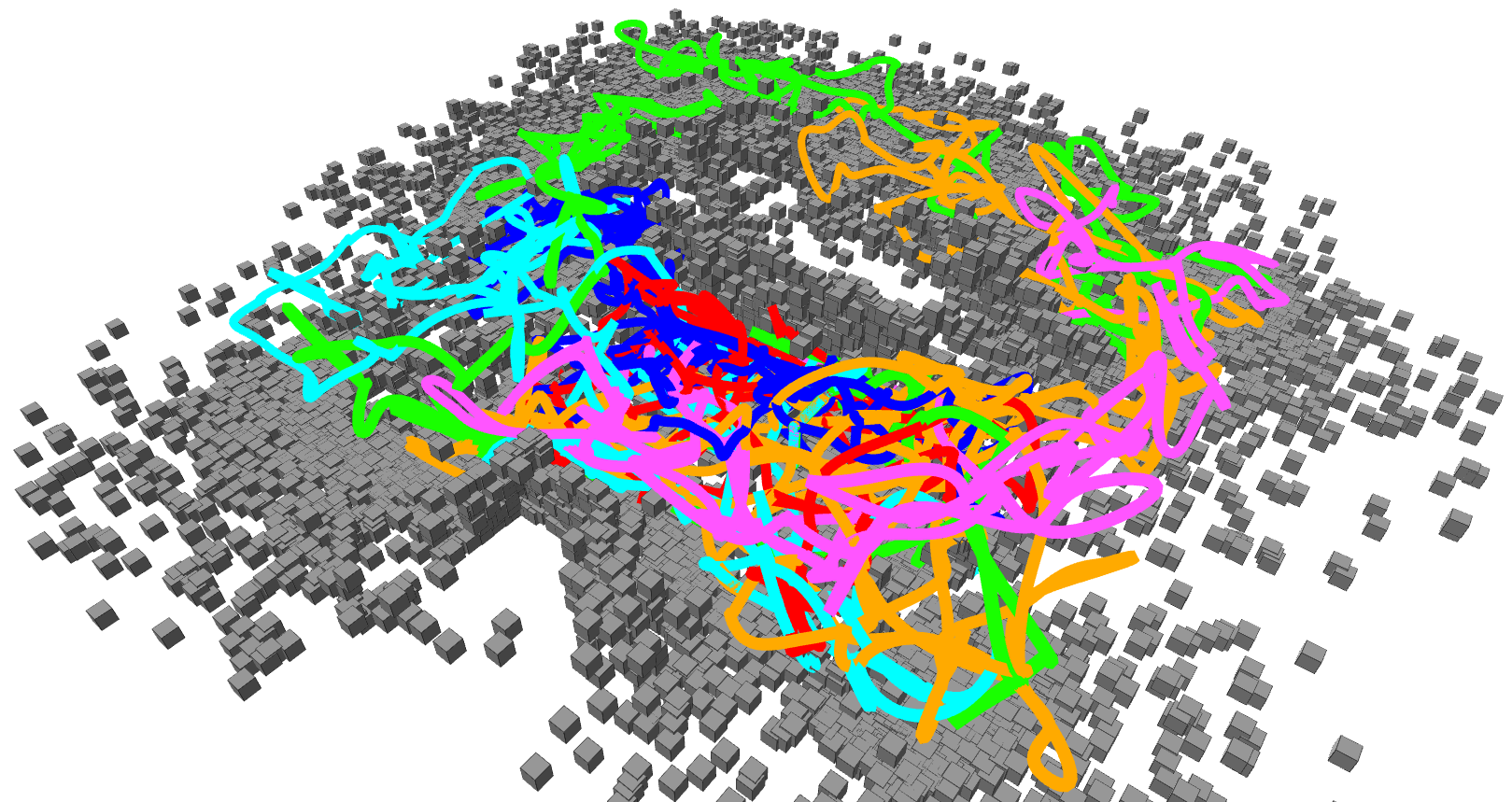}};
  \end{tikzpicture}
\end{minipage}
  \centering
  \caption{{Ablation test results on the earthquake world. Left: visualization of exploration progress, best runs per method are thick and a marker signifies a crash or mission completion. Right: best-run trajectories for each method.}
  }
  \label{fig:ablations}

\end{figure}

\begin{table}[h!]
  \caption{{Quantitative results showing the explored area in the simulation experiments described in \autoref{sec:sim_exper}.} }
\centering
  \begin{tabular}{m{18mm} m{12mm} m{10mm} m{10mm} m{10mm}} 
 \hline
    \textbf{Method / Ablation} & \textbf{World} & \textbf{Mean} $A \left[m^2 \right] $ & \textbf{Max} $A \left[ m^2 \right] $ & \textbf{Max} $V \left[ m^3 \right] $\\ [0.5ex] 
 \hline\hline
    MonoSpheres  & Earthquake & \textbf{1968}& \textbf{2432}& \textbf{17706} \\
    No FFF  & Earthquake & 988& 1272& 9518 \\
    No DBOF  & Earthquake & 923& 1232& 10356 \\
    No OVDE  & Earthquake & 299& 345& 1837 \\
    No VPB  & Earthquake & 1479& 1670& 13475 \\
    Grid-Based & Earthquake  & 764& 922& 5181 \\
    \hline
    MonoSpheres  & Cave & \textbf{3051}& \textbf{3202} & \textbf{18350} \\
    Grid-Based & Cave  & 1268& 1350& 8794 \\
    \hline
    MonoSpheres  & Rooftops & \textbf{1504}& \textbf{1842}& \textbf{21713} \\
    Grid-Based & Rooftops  & 544& 622& 3975 \\
    \hline
    MonoSpheres  & Sparse & \textbf{2051}& \textbf{2060}& \textbf{14193} \\
    Grid-Based & Sparse  & 843& 867& 3413 \\
 \hline
\end{tabular}
\label{table:sim_exper}
\end{table}

\subsection{Multi-World Simulation Evaluation}
\label{sec:sim_exper}
As the previously published monocular exploration methods do not have code available (with the exception of \cite{simon2023mononav}, which, however, is not ROS-compatible as of writing this paper), we prepared a simple grid-based mapping and exploration pipeline for comparison with MonoSpheres.
This reference method uses the same mapping approach as \cite{from_monoslam_to_explo, los_maps} (i.e. initializing free space between the camera and the visual SLAM points that have a low position covariance). 
{In the following text, we show that it is the usage of this classical mapping method that limits exploration with sparse depth data in these previous works.
To explore, the method periodically samples free-space points uniformly in a 40x40x10 \SI{}{\meter} area around the robot and sends these as goals to a grid-based A* path planning module, which finds path through the occupancy grid with the same minimum distance from obstacles and \fix{unknown} space as the MonoSpheres method (\SI{1.5}{\meter} in these experiments).
The robot then follows these paths with the camera pointing forward and the planner replans at \SI{5}{\hertz} to handle measurement \fix{inaccuracies}.}
We use OctoMap \cite{octomap} for the grid-based mapping with a \SI{0.5}{\meter} cell size.
This value was chosen because smaller cell sizes caused the robot to not find any safe paths due to the free-space gaps documented in \cite{from_monoslam_to_explo} and larger cell sizes blocked exploration of narrow areas. 

As the evaluation metric, the volume of the constructed map is commonly used in exploration literature.
However, because of the open-area free-space sampling scheme and the depth interpolation, MonoSpheres constructs more free space than the raytracing mapping approach over the same trajectories. 
{For this reason, the volume metric is not fair towards methods using the grid-based raytracing mapping and we focus more on explored \textit{area}.
We divide the target exploration area into 2.5x2.5m columns that are marked as explored if the constructed map (sphere-based or grid-based) contains any element in a given column.}
The metric, $A$, is then the top-down 2D area of all the explored columns.

{We compare the classical mapping + random exploration approach with MonoSpheres with fixed parameters in simulations on 4 different worlds in 3 runs per method (6 in the Earthquake world for better comparison), and terminate each run after \SI{20}{\minute} (\SI{40}{\minute} in the Cave world due to its size) or sooner if a collision occurs.
The resulting maps and trajectories are shown in \autoref{fig:exper_fireworld} and the explored area and volume in \autoref{table:sim_exper}.}
Videos from example runs are available in the multimedia materials.
The experiments were conducted in the Gazebo simulator \cite{gazebo} with the same UAV platform and sensory setup as in the real-world experiments.
{Exploration goal generation was constrained by a 3D bounding box in each outdoor world for both methods.}

\fix{We designed the Earthquake environment to test the limits of MonoSpheres with the OpenVINS SLAM frontend, making it contain textureless surfaces and thin obstacles.
As expected, this has led to collisions in both MonoSpheres and the grid-based mapping method (see \autoref{fig:ablations}), as these objects cannot be detected by the sparse visual SLAM frontend.
In particular, the grid-based method collided once with a smooth metal rod, and the MonoSpheres method collided with the textureless roof of a bus.}

\subsection{Ablation Experiments}
\label{sec:ablations}
To determine the contribution of the individual components of our proposed mapping and exploration method, we performed ablation tests in the earthquake world with 3 exploration runs for each ablation.
The results are shown in \autoref{table:sim_exper} and \autoref{fig:ablations}.
Without DBOF, the UAV tends to quickly overwrite previously mapped obstacles with free space and crash.
Without forward-facing flight (FFF), where the UAV aligns itself with a goal viewpoint's heading immediately after starting to navigate to it, the UAV is unable to detect previously undetected obstacles and also crashes quickly.
Without virtual depth (OVDE), the UAV is unable to explore open-space areas and stays in the starting "valley" between the houses as expected.
Since the grid-based method allows casting thin rays even to faraway points, it maps a larger area than this ablation.
Lastly, removing viewpoint blocking (VPB) does not lead to critical failures, as the simulated environment does not have large featureless walls, but the exploration performance is still notably reduced due to the UAV revisiting some viewpoints.
In summary, these results confirm the importance of all the proposed components of our mapping and exploration approach.

\section{\sectionfix{Discussion and Future Work}}
\label{sec:discussion}
{

\fix{
The results of the simulations and real-world experiments show that MonoSpheres is capable of exploring many different topologies of environments --- sparse, dense, indoor, outdoor and environments where flight over obstacles in wide open spaces is required.
As expected based on the findings of \cite{from_monoslam_to_explo, los_maps}, the classical mapping approach with random exploration goal sampling was only able to map the area close to tall obstacles near the starting position.
This, along with ablation results, supports the necessity of the OVDE module for mapping open spaces, which in turn requires the DBOF to avoid overwriting previously mapped obstacles.

The main limitation of MonoSpheres lies in the detection capabilities of the visual SLAM frontend used for depth estimation --- the mapping module cannot map obstacles that the frontend does not detect (thin objects or large textureless objects).
This could be solved in future work by integrating low-cost but noisy short-range ultrasonic sensors into the mapping pipeline, 
or by switching to a dense SLAM frontend (at the cost of higher computational complexity).
}

}

\section{CONCLUSION}
{In this paper, we presented a novel, perception-coupled approach to mapping and planning for monocular UAV 3D exploration.
Through ablation experiments, we have confirmed the importance of the individual modules of our proposed method that are tightly coupled to the properties of monocular motion-based depth sensing.
Compared to the existing approaches to 3D monocular exploration, which have so far presented autonomy in highly specific environments and at the scale of a few indoor rooms \cite{from_monoslam_to_explo, los_maps, simon2023mononav}, 
\fix{MonoSpheres} achieves autonomous exploration at considerably larger scales, both in simulation and in the real world, in a diverse set of environments.}










\bibliographystyle{IEEEtran}
\bibliography{main.bib}

\end{document}